\documentclass[twocolumn]{cinc}
\usepackage{graphicx}

\begin{document}
\bibliographystyle{cinc}

\title{Deep Learning for End-to-End Atrial Fibrillation Recurrence Estimation}


\author { Riddhish Bhalodia$^1$, Anupama Goparaju$^1$, Tim Sodergren$^1$, Alan Morris$^2$, Evgueni Kholmovski$^{2,3}$,\\
	Nassir Marrouche$^2$, Joshua Cates$^2$, Ross Whitaker$^1$, Shireen Elhabian$^1$ \\
\ \\ 
$^1$ Scientific Computing and Imaging Institute, School of Computing, \\ 
University of Utah, Salt Lake City, Utah, USA
 \\
$^2$ Comprehensive Arrhythmia Research and Management Center, Division of
 Cardiovascular Medicine, School of Medicine, University of Utah, Salt Lake City, Utah, USA
\\
$^3$ Department of Radiology and Imaging Sciences, School of Medicine, \\ University of Utah, Salt Lake City, Utah, USA
 } 
\maketitle

\begin{abstract}
Left atrium shape has been shown to be an independent predictor of recurrence after atrial fibrillation (AF) ablation. Shape-based representation is imperative to such an estimation process, where correspondence-based representation offers the most flexibility and ease-of-computation for population-level shape statistics.
%
%
Nonetheless, population-level shape representations in the form of image segmentation and correspondence models derived from cardiac MRI require significant human resources with sufficient anatomy-specific expertise. 
In this paper, we propose a machine learning approach that uses deep networks to estimate AF recurrence by predicting shape descriptors directly from MRI images, with NO image pre-processing involved. 
%
%
We also propose a novel data augmentation scheme to effectively train a deep network in a limited training data setting. 
%
%
We compare this new method of estimating shape descriptors from images with the state-of-the-art correspondence-based shape modeling that requires image segmentation and correspondence optimization. 
Results show that the proposed method and the current state-of-the-art produce statistically similar outcomes on AF recurrence, eliminating the need for expensive pre-processing pipelines and associated  human labor. 
\end{abstract}
\vspace{-0.3in}
\section{Introduction}
\vspace{-0.1in}
Left atrium (LA) shape has been shown to be associated with atrial fibrillation (AF) \cite{marrouche2014decaaf, bisbal2013sphericity}. In the past three decades, catheter ablation has evolved to a safe, effective, and clinically acceptable therapeutic option in patients with AF, and is recognized as the first-line therapy in patients with symptomatic paroxysmal and persistent AF \cite{calkins20172017}, but with considerable post-ablation recurrence leading to high rates of redo ablations that reach to 40\% \cite{calkins20172017, darby2016recurrent}.
%
Recent studies demonstrate the efficacy of using LA shape as a predictor for AF recurrence after catheter ablation \cite{bieging2018left,cates2014computational,cates2013afib}.
However, such studies rely heavily on patient-specific LA shape representation that entails a time-consuming, expert-driven, expensive, irreproducible, and error-prone workflow of segmenting patient's LA from cardiac MRI volumes followed by a processing pipeline of shape registration and dense placement of corresponding landmarks, which requires significant amount of human resources and anatomy-specific expertise. 

Landmarks are the most popular choice as a light-weight and general shape representation suitable for statistical analysis and visual communication of the results \cite{sarkalkan2014statistical}. Landmarks should be defined consistently within a population to refer to the same, i.e., \textit{corresponding}, anatomical position on every shape instance,  resulting in correspondence-based models ---aka point distribution models (PDMs). Such models and derived population-level statistics play an important role in ablation guidance, fibrosis quantification, and biophysical modeling in AF patients \cite{calkins20172017,marrouche2014decaaf,krueger2013silico}. Furthermore, these models are useful in deriving shape-driven LA segmentation methods to account for misleading imaging information \cite{tobon2015benchmark}. Nonetheless, PDM generation state-of-the-art methods (e.g., \cite{cates2017shapeworks,durrleman2014morphometry,styner2006spharm}) rely on manual segmentations and a subsequent heavy pre-processing of shape registration and correspondence optimization (and associated algorithmic parameters tuning).

Recently, deep networks have found many uses in medical image analysis. Their natural ability to learn complex functions makes them ideal for more complex computer-aided tasks such as automatic segmentation \cite{ronneberger2015unet} and landmark detection \cite{zheng2015detection}. In this paper, we propose an automated approach that rely on deep networks to generate a patient-specific landmark-based anatomy representation directly from 3D cardiac MRI, hence, negating any need for manual pre-processing and segmentation. 
%
However, a deep network cannot be viably trained with limited training samples ---a typical situation in this application and in many similar medical imaging applications.
To mitigate this problem, we propose a novel data augmentation scheme that generates more statistically feasible data, and hence enable training the deep network while reducing the associated risk of overfitting.
%
We compare the efficacy of the proposed method with the state-of-the-art correspondence-based shape modeling, which requires segmentation and correspondence optimization, in terms of correspondence-level error and difference in AF recurrence prediction. Results show statistical equivalence. As the proposed method is based on learning shape descriptors from images, hence, the method has been leveraged for automatic LA segmentation with  promising results.
\vspace{-0.1in}
\section{Methods}
\begin{figure}[!h]
\centering
\includegraphics[width=0.4\textwidth]{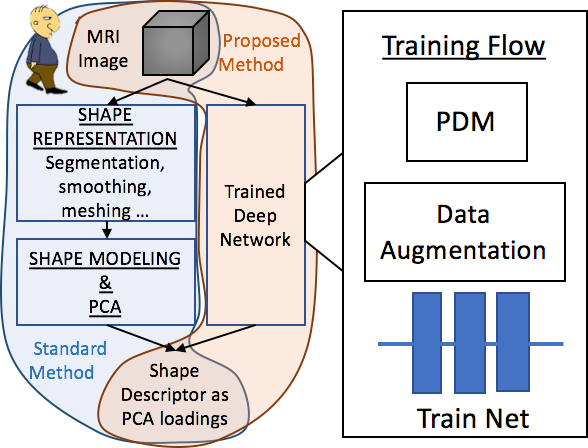}
\vspace{-0.1in}
\caption{A pictorial representation of the existing standard shape modeling pipeline and the proposed pipeline. The blue-shaded pipeline requires manual pre-processing and segmentation for every new data to be processed. In contrast, once the network is trained, the proposed method (shaded in orange) is fully automatic.}
\vspace{-0.2in}
\label{fig:trainmodel}
\end{figure}

 \vspace{-0.1in}
Figure \ref{fig:trainmodel} shows the workflow of the proposed method, as compared to the standard workflow, including model training and the usage of the trained network for new images.

\begin{figure}
\centering
\includegraphics[scale=0.38]{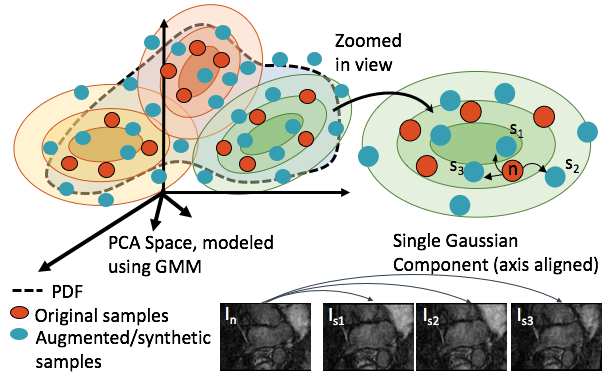}
\vspace{-0.2in}
\caption{Pictorial representation of the data augmentation scheme, we use GMM to find pdf in PCA subspace and then use each Gaussian component to generate new samples.}
\vspace{-0.2in}
\label{fig:data-augmentation}	
\end{figure}

\vspace{0.05in}
\noindent\textbf{Data augmentation: }
We use a population of 207 MRI scans of LA of AF patients (\textit{original data}), and these are not enough to train a deep network with, given the high dimensionality of such images (and their respective LA shapes).
We propose the following data augmentation scheme. We first compute the PDM of the original data using \emph{ShapeWorks} \cite{cates2017shapeworks} software. Then, we perform principal components analysis (PCA) on the original data, which reduces the dimension of each sample from thousands of 3D points to 15 PCA loadings (dominant modes that capture 95\% of shape variability). This PCA subspace represents a multivariate Gaussian that describes the shape variations in the given population. Further, this subspace parameterizes the data distribution, enabling generating thousands \textit{shape} samples that respect the population-level statistics. To obtain the corresponding MRI \textit{image} of a generated shape sample, we find the nearest neighbor to the generated sample from the original data and use the correspondences as landmarks, to perform thin plate spline (TPS) warp \cite{bookstein1989principal}. 
%
%
The LA structure exhibits a natural clustering in the \emph{shape space} due to the variability of pulmonary veins arrangements  \cite{sohns2011mdct}. To account for such variability, we model LA shapes as a multi-modal Gaussian distribution in the PCA subspace, with 3 components yielding the best Bayesian information criterion (BIC). The method is pictorially depicted in Figure \ref{fig:data-augmentation}.
In our experiments, we used 175 samples out of 207 to use for data augmentation, and the rest is set aside for deep network testing (as \textit{unseen} samples). 

\vspace{0.05in}
\noindent\textbf{Network architecture and training: }
To predict/estimate the shape descriptor in the form of 15 PCA loadings directly from the 3D MRI scans, we use a deep convolution neural network (CNN) \cite{lecun1998cnn} that uses the L2 loss function for optimization. We train the network for 240 epochs using Adagrad optimizer in Tensorflow. We use 5000 training samples generated from the three mixture components as described in the data augmentation scheme. 


\vspace{0.05in}
\noindent\textbf{AF prediction: }
AF recurrence has been hypothesized to be dependent on shape descriptors such PCA loadings \cite{marrouche2014decaaf}. We hypothesize that the PCA loadings predicted using the proposed deep network are equivalent to the ones from the state-of-the-art pipeline with regards to AF recurrence prediction. Hence, we train a multi-layer perceptron (MLP) using the PCA loadings of the original 175 data derived from \emph{ShapeWorks} correspondences, and their AF recurrence data (binary variable). The trained MLP predicts the AF recurrence probability that can be compared using the PCA loadings computed using the deep network and through the standard state-of-the-art PDM model. 

\vspace{-0.1in}
\section{Results}
\vspace{-0.1in}

We use the PDM on 207 MRI's as our base data, of which 175 are used for data augmentation and hence, influenced the network training. The remaining 32 are called the \emph{cross-validation} data that is used for several validation scenarios as detailed in the following.

\vspace{0.05in}
\noindent\textbf{PCA loadings and correspondence estimation: }
To serve as an automated surrogate to the standard workflow, the proposed method needs to produce valid patient-specific correspondence model. We validate this assessment using three different aspects. The network produces the output in form of PCA loadings, hence, we first compare these estimated PCA loadings with those given by the PDM, i.e., the ground truth. As these are multivariate data, we use Hotelling $T^2$ statistic to identify if the differences were statistically significant. We obtain the statistic to be 11.1 with 78\% confidence (97\% for 175 data used for training and 74\% for the data used for cross-validation). 

\begin{figure}[!t]
\centering
\includegraphics[scale=0.6]{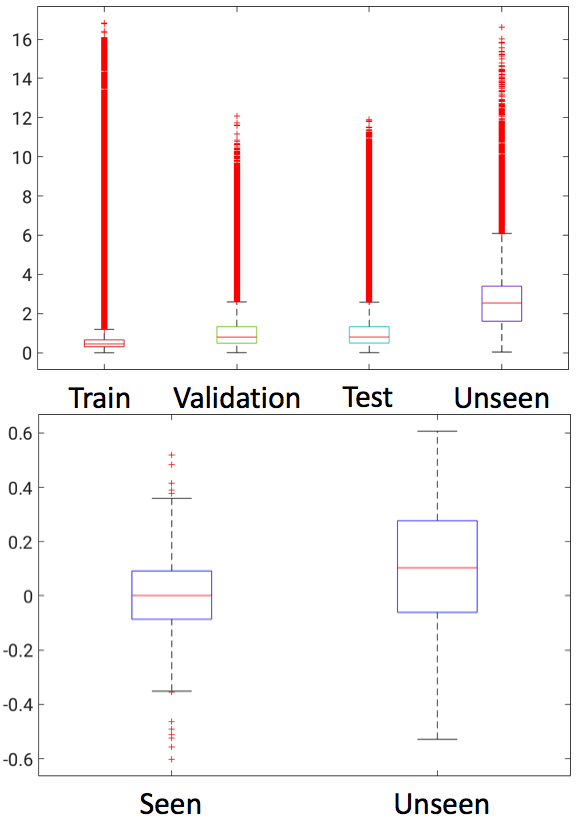}
\vspace{-0.1in}
\caption{(Top) Box plots of per-point per-shape error  in \emph{millimeters} between the reconstructed correspondences produced through deep network, and, the ground truth correspondences. (Bottom) Difference between the recurrence probabilities from ground truth PCA loadings and the network produced PCA loadings, ``seen" data is the original data used for PDM generation and ``unseen" is again the cross-validation data.}
\vspace{-0.2in}
\label{fig:boxplots}
\end{figure}

To evaluate the efficacy of the network to provide loadings that correctly reconstruct the patient-specific correspondence model, we reconstruct the correspondences produced through the network and compare it with the ground truth correspondences. Boxplots in Figure \ref{fig:boxplots} show the per-point per-shape error in millimeters (\textit{mm}) for all training, validation, testing and cross-validation data (called ``unseen" in the figure). The errors for all but the unseen data are less than the voxel spacing of the images (2 \textit{mm})  in average, and thereby achieving sub-voxel accuracy.

\begin{figure}[!h]
	\centering
	\includegraphics[scale=0.6]{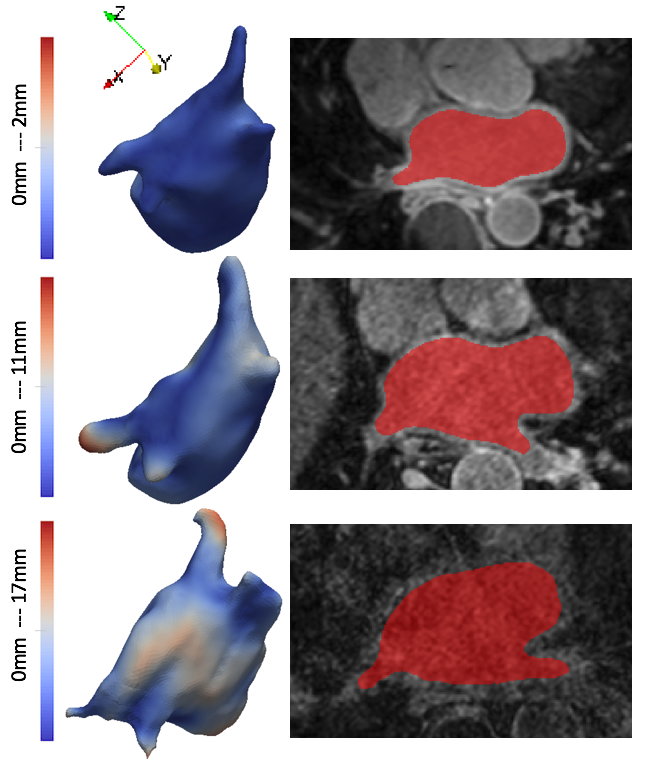}
	\vspace{-0.1in}
	\caption{This figure shows three different scans from the cross-validation data. Each row has the image and it's ground truth segmentation, and it also has a mesh surface with a surface to surface error heatmap (between ground truth segmentations and the reconstructed meshes using the network) overlay. From top to bottom we see images of degrading quality as well as the drastic shape differences in the LA surface meshes.}
	\label{fig:hausdaurff}
	\vspace{-0.2in}
\end{figure}

To evaluate the use of the proposed method as a possible automatic segmentation method or as an approximate driver towards LA segmentation, we use the meshes generated from the reconstructed correspondences, and we compare them with the original ground truth segmentations. For the validation metric, we use surface-to-surface distance between the reconstructed meshes and the original ones. In Figure \ref{fig:hausdaurff}, we observe that the largest surface-to-surface distances are concentrated in the pulmonary veins, but the piece-wise continuous regions of the LA anatomy are reconstruction with relatively minimal error. We also observe the degrading image quality from top to bottom in Figure \ref{fig:hausdaurff} and the increase in the associated error. It is worth emphasizing that no pre-processing is performed when feeding these images to train the deep network.

\vspace{0.05in}
\noindent\textbf{AF recurrence prediction: }
The deep network should estimate shape parameters that subsequently can predict accurate outcome analysis (AF recurrence prediction in our case). 
To assess this aspect, we test for equivalence between the AF recurrence probabilities predicted by the PCA loadings from the PDM and that from the deep network using the two one-sided test (TOST) \cite{schuirmann1987tost}. We find the recurrence predicted by deep network and ground truth PCA loadings to be equivalent with a confidence of 90\% with the mean difference bounds of $\pm 0.06$. We also report the error between both predictions in a boxplot as shown in Figure \ref{fig:boxplots}. 

\vspace{-0.1in}
\section{Conclusion}
\vspace{-0.1in}
We proposed a pre-processing-free method for LA shape analysis and demonstrated its usability in predicting AF recurrence. We compared this method with the standard state-of-the-art shape analysis workflow, which requires significant human supervision and anatomy-specific expertise, and found the results to be statistically comparable. The proposed method directly uses images and it is also shown to be a viable option for automatic LA segmentation. Errors of the method can be attributed to the huge variability in image intensities arising from different scanners and acquisition protocols. We believe that using an improved data augmentation method that takes into account the intensity variability and shape statistics of the data will improve the network generalization.

\vspace{0.03in}
\noindent\textbf{Acknowledgements: } This work was supported by the National Institutes of Health [grant numbers R01-HL135568-01 and P41-GM103545-19]. 
\vspace{-0.1in}
\bibliography{refs}
\vspace{-0.2in}







\begin{correspondence}
Riddhish Bhalodia\\
72 Central Campus Drive, 3rd floor desk, Salt Lake City, Utah-84112, USA  \\
riddhishb@sci.utah.edu
\end{correspondence}

\end{document}